\documentclass[conference]{IEEEtran}

\makeatletter
\def\ps@IEEEtitlepagestyle{%
  \def\@oddfoot{\mycopyrightnotice}%
  \def\@evenfoot{}%
}
\def\mycopyrightnotice{%
  {\footnotesize 978-1-6654-7095-7/22/\$31.00~\copyright~2022 IEEE\hfill}
  \gdef\mycopyrightnotice{}
}

\usepackage{blindtext}
\usepackage{eso-pic}
\IEEEoverridecommandlockouts
\usepackage{cite}
\usepackage{amsmath,amssymb,amsfonts}
\usepackage{algorithmic}
\usepackage{graphicx}
\usepackage{textcomp}
\usepackage{xcolor}
\usepackage{subcaption}
\usepackage{url} 
\def\BibTeX{{\rm B\kern-.05em{\sc i\kern-.025em b}\kern-.08em
    T\kern-.1667em\lower.7ex\hbox{E}\kern-.125emX}}
    
\usepackage{eso-pic}
\newcommand\AtPageUpperMyright[1]{\AtPageUpperLeft{%
 \put(\LenToUnit{0.17\paperwidth},\LenToUnit{-2cm}){%
     \parbox{0.9\textwidth}{\raggedleft\fontsize{8}{11}\selectfont #1}}%
 }}%
\newcommand{\conf}[1]{%
\AddToShipoutPictureBG*{%
\AtPageUpperMyright{#1}
}
}

\begin{document}
\title{\vspace*{1cm} Suppress with a Patch: Revisiting Universal Adversarial Patch Attacks against Object Detection\\
 }

\author{\IEEEauthorblockN{Svetlana Pavlitskaya}
\IEEEauthorblockA{
\textit{FZI Research  Center  for  Information  Technology}\\
Karlsruhe, Germany\\
pavlitskaya@fzi.de}
\and
\IEEEauthorblockN{Jonas Hendl}
\IEEEauthorblockA{\textit{Karlsruhe Institute of Technology}\\
Karlsruhe, Germany \\
 jonas.hendl@student.kit.edu}
\and
\IEEEauthorblockN{Sebastian Kleim}
\IEEEauthorblockA{\textit{Karlsruhe Institute of Technology}\\
Karlsruhe, Germany \\
 sebastian.kleim@student.kit.edu}
\and
\IEEEauthorblockN{Leopold Johann Müller}
\IEEEauthorblockA{\textit{Karlsruhe Institute of Technology}\\
Karlsruhe, Germany \\
 leopold.mueller@student.kit.edu}
\and
\IEEEauthorblockN{Fabian Wylczoch}
\IEEEauthorblockA{\textit{Karlsruhe Institute of Technology}\\
Karlsruhe, Germany \\
 uermo@student.kit.edu}
\and
\IEEEauthorblockN{J. Marius Z\"ollner}
\IEEEauthorblockA{\textit{FZI Research  Center  for  Information  Technology}\\
\textit{Karlsruhe Institute of Technology}\\
Karlsruhe, Germany \\
zoellner@fzi.de}
}
\maketitle
\conf{\textit{  Proc. of the International Conference on Electrical, Computer, Communications and Mechatronics Engineering  (ICECCME) \\ 
16-18 November 2022, Maldives}}
\begin{abstract}
Adversarial patch-based attacks aim to fool a neural network with an intentionally  generated noise, which is concentrated in a particular region of an input image. In this work, we perform an in-depth analysis of different patch generation parameters, including initialization, patch size, and especially positioning a patch in an image during training. We focus on the object vanishing attack and run experiments with YOLOv3 as a model under attack in a white-box setting and use images from the COCO dataset. Our experiments have shown, that inserting a patch inside a window of increasing size during training leads to a significant increase in attack strength compared to a fixed position. The best results were obtained when a patch was positioned randomly during training, while patch position additionally varied within a batch.
\end{abstract}

\section{Introduction}\label{introduction}

Deep neural networks have been shown to possess an inherent vulnerability to adversarial inputs, i.e. deliberately generated noise patterns, which drastically change model prediction~\cite{szegedy2013intriguing,goodfellow2014explaining}. For images, an adversary can either generate small imperceptible pixel changes over the whole image or concentrate the visible noise in a specific image region, named \textit{adversarial patch}~\cite{brown2017adversarial}. Adversarial patches are especially dangerous because they can be printed to perform attacks in the real world.

In this work, we focus on the vanishing attack on object detection and analyze the hyperparameters involved in the generation of universal adversarial patches. In particular, we evaluate the impact of the patch size and the initialization approach. To achieve location-agnostic patches, we evaluate three possible patch placement strategies: placing a patch at a fixed position, dynamic window approach, and random patch placement. We furthermore assess the impact of intra-batch variation of patch positions. Our study thus provides suggestions for adjusting the training procedure in order to generate a location invariant patch.
Differently from the existing work, we compute the attack success rate by comparing a manipulated image with the one containing a benign white patch, and not with a clean image. It ensures that a possible impact of simple overlapping is not ignored when patch attack strength is evaluated.

A large portion of the existing work focuses on placing a patch directly onto an object, with a goal of suppressing only this object~\cite{song2018physical,thys2019fooling,wu2020making}. Our approach is closer to that of Lee and Kolter~\cite{lee2019physical}, because we do not require a patch to be placed directly on an object and aim to suppress all object detections in input image. This setting allows for more realistic and dangerous real-world attacks, e.g. in a self driving scenario.

\begin{figure}[t]
\centering
\begin{subfigure}[t]{0.485\linewidth}
    \includegraphics[width=1.0\textwidth]{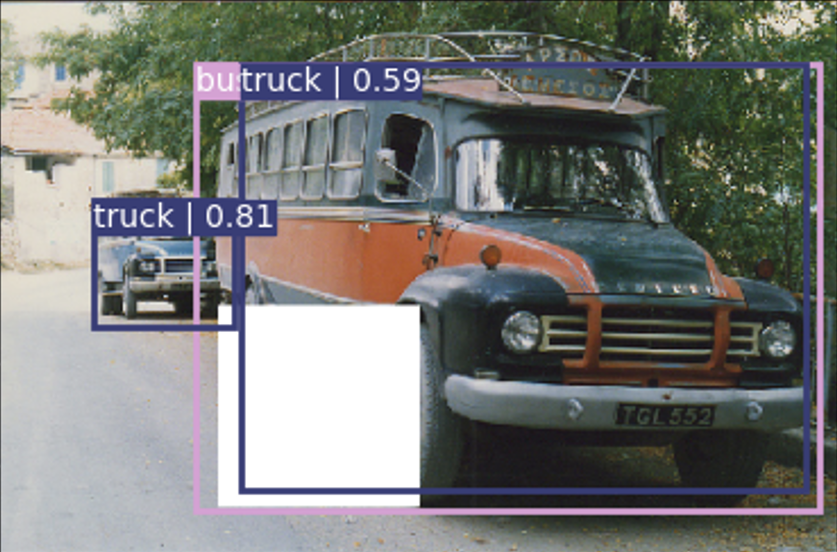}
  	\caption{White patch}
\end{subfigure}
\begin{subfigure}[t]{0.485\linewidth}
  \includegraphics[width=1.0\textwidth]{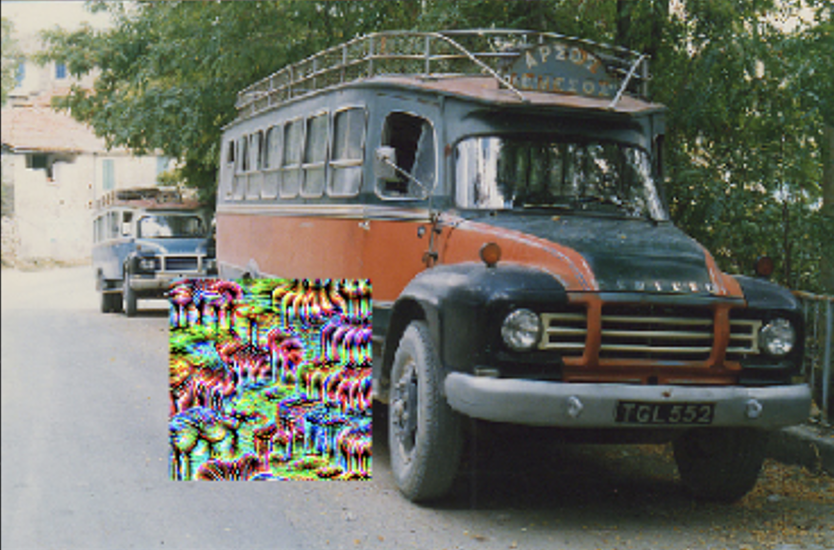}
  	\caption{Adversarial patch}
\end{subfigure}
\caption{We evaluate universal adversarial patch attacks against object detection. The attack goal is to suppress all objects in the input. Note, that the effect is not caused by overlapping, because white patch does not interfere with the network prediction.}
    \label{fig:attackallobjects}
\end{figure}

\section{Related work}\label{relatedwork}

Adversarial attacks, initially discovered by Szegedy et al. \cite{szegedy2013intriguing}, consist in adding intentionally generated yet invisible noise to an image, thus causing a neural network to come to a wrong prediction with high confidence. The seminal work by Brown et al. \cite{brown2017adversarial} has first introduced locally-bound yet visible adversarial noise, i.e. an adversarial patch. Since then, adversarial patch-based attacks against various computer vision tasks have been proposed~\cite{karmon2018laVAN,nesti2022evaluating,pavlitskaya2020feasibility}. 

The DPatch approach by Liu et al.\cite{liu2018dpatch} extended the original method by Brown et al. to the object detection task. The optimization goal was to maximize error of the bounding box regression and object classification simultaneously. For the targeted scenario, the attack success for different classes was analyzed and the most efficient classes for attack were found. DPatch deliberately avoided overlapping with the objects by placing a patch at a fixed position onto the top left corner of the image. DPatch experiments with YOLO and  Faster R-CNN, however,  remained within the digital setting. Lee and Kolter~\cite{lee2019physical} have focused on making a patch attack feasible in the real world. In particular, they have applied  the Expectation over Transformations (EoT) approach by Athalye et al.~\cite{athalye2018synthesizing} to the patch, thus ensuring rotation, scaling, and translation invariance as well as robustness to brightness changes. Maximizing training loss for the ground truth labels was used to achieve suppression of object detections.  Their experiments with YOLOv3 both in digital and realistic settings have demonstrated significant detection accuracy drop. For the real-world evaluation, a printed version of the patch was used.

The feasibility of adversarial patch attacks in the real world was first approached by Song et al.~\cite{song2018physical}. This work focused on attacking stop sign detection performed with YOLOv2. Two types of attacks were demonstrated: the disappearance attack made a stop sign invisible, whereas the creation attack made the attacked model detect non-existent objects. To achieve it, the RP2 (Robust Physical Perturbations) approach ~\cite{eykholt2018robust} was extended to the object detection case, and a new adversarial loss was introduced. A close work by Chen et al.~\cite{chen2018shapeshifter} attacked Faster R-CNN for stop sign detection. Their approach relied on EoT~\cite{athalye2018synthesizing} to achieve strong attack under different view angles in drive-by scenarios.

More recent works aimed at realistic attacks for more challenging object classes and beyond restricted scenes. For example, Thys et al.~\cite{thys2019fooling} focused on suppressing persons via an overlapping patch. Due to the high level of intra-class variance, the attack was especially challenging and required using a specific adversarial loss. Later works have also experimented with printing an adversarial patch on a t-shirt~\cite{xu2020adversarial,wu2020making}.

Recently, an effort has been made to ensure naturalistic appearance of the adversarial patches to make them less conspicuous for humans and thus cause more dangerous attacks~\cite{hu2021naturalistic,pavlitskaya2022feasibility}.

\section{Attack Scenario}\label{methodology}

Our attack pipeline relies on the projected gradient descent (PGD) algorithm~\cite{madry2017towards}, all attacks are performed in a universal manner \cite{moosavi2017universal}, i.e. a single patch is generated to attack all possible input data. We focus on the object vanishing attack~\cite{chow2020targeted}, thus aiming to suppress all object detections in an input image.

To perform an attack, we insert a patch into an  image by replacing the corresponding image pixels with patch pixels. To asses the impact of the patch size, we consider the \textit{coverage}, i.e. the ratio of the patch size to the patch image.

The loss function of the attacked object detector consists of three components: the classification loss, the localization loss, and the confidence loss (i.e. the objectness of the detection)~\cite{redmon2018yolov3}. Our preliminary experiments have demonstrated, that the latter loss component has the largest impact on attack success. Therefore, the loss function that we use for attack relies on the objectness score. Suppose we obtain the objectness scores $O_{l,p}$ for each detection $d$ of each layer $l$. We can then define the loss function to be minimized with respect to the patch \(P\) as follows:

\begin{equation}\label{lossfunction}
   Loss_{Attack} = \underset{P}{argmin} \sum^{3}_l \frac{\sum^P_p ReLU(O_{l,d})}{D}
\end{equation}

For attacks in the real world, a patch should be robust to different positions in input and also different patch sizes. The simplest implementation is to insert a patch at the center of an image as a naive yet unrealistic benchmark. In order to ensure translation robustness, a patch is inserted at a random position. Alternatively, we propose the \textit{dynamic window} approach -- we define a window in which a patch can appear and increase its size during training. Finally, we suggest additional variation of the patch position not only between batches but also within a batch to ensure location invariance. 

To evaluate our results and compare different patches, we introduce the \textit{Attack Success Rate (ASR)} metric, defined as follows:

\begin{equation}\label{attacksuccessrate}
    ASR =\frac{\sum^N_{i=1}max(0,1-\frac{N_{adv\_objects}}{N_{dummy\_objects}})}{N},
\end{equation}

where $N$ is the number of images used for the respective run, $N_{adv\_objects}$ is the number of objects detected in an image modified with an adversarial patch, and $N_{dummy\_objects}$ the number of detected objects in the same image but instead with a plain white patch to take covered objects into account. The ASR takes values between 0 and 1, while 1 means that the object detector is no longer able to detect any objects in the input. Replacing white with a different color for patches used in comparison has revealed no significant difference.

\section{Experiments and Results}\label{ablationexperiments}

\subsection{Experimental Setup}\label{experimentalsetup}

For the experiments, we use the COCO dataset~\cite{lin2015microsoft} and the open source PyTorch implementation of YOLOv3\footnote{Available at \url{https://github.com/eriklindernoren/PyTorch-YOLOv3}}~\cite{redmon2018yolov3}, which uses the Darknet-53 weights. We set the confidence threshold to $0.5$ and the intersection over union threshold to $0.4$. The learning rate is set to $0.1$, lower learning rates have demonstrated no advantage in the training process. The images are processed at resolution 416x416.

Patches were trained using PGD with Adam optimizer on the COCO training subset and evaluated by inserting them into the testing subset images. For evaluation, we used the COCO validation data set from 2017 containing 5,000 images. All trainings were performed on an NVIDIA RTX 1080 Ti GPU with 11GB VRAM.

\subsection{Patch Size}\label{sizeofpatch}
To assess the effect of different patch sizes, we run an evaluation on a smaller subset of eight test images. 

Larger patches evidently provide larger attack surface and thus lead to more successful attacks. However, larger patches are less plausible for real-world attack scenarios, since they need to be either close to the camera or physically very large which makes printing and carrying them unfeasible. Furthermore, a larger patch also tends to cover objects to a larger extent. The proposed ASR metric, however, takes it into account via comparison to the corresponding white patches of the same size and location.

\begin{figure}[t]
    \centering
    \includegraphics[width = \linewidth]{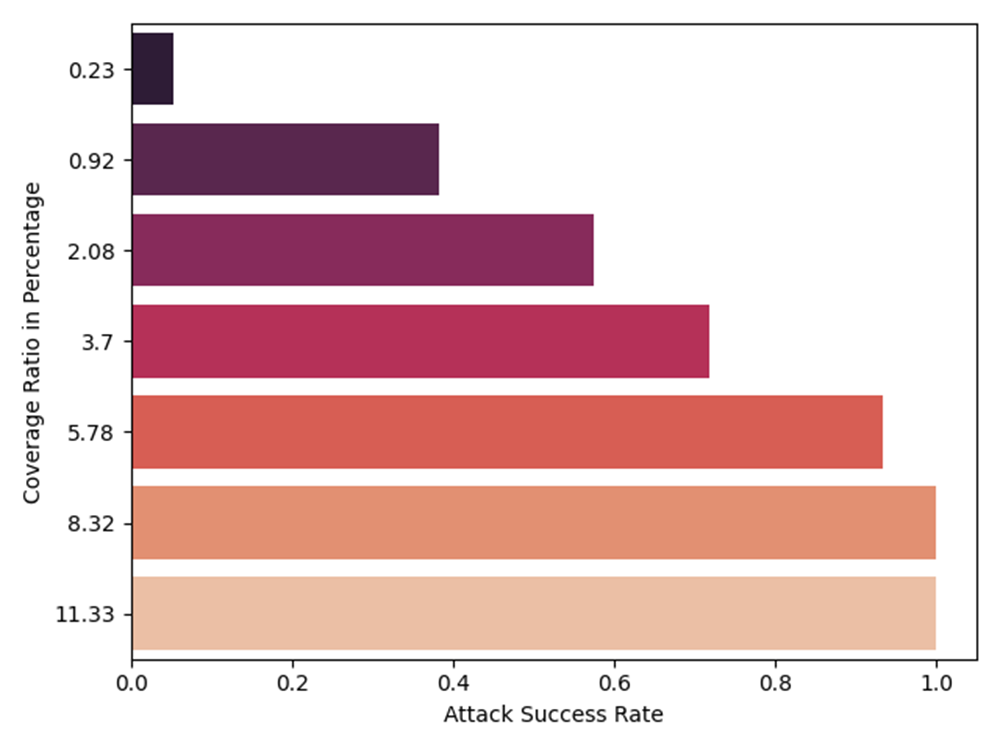}
    \caption{ASR for patches of different size. The smallest patch size 20x20 pixels, the size further increases with a step of 20 pixels, the largest evaluated patch size is 140x140 pixels. Coverage denotes the ratio of image pixels, occupied by a patch, to the overall number of pixels in an image.}
    \label{fig:patchsizevisual}
\end{figure}

\begin{figure}[t]
    \centering
    \includegraphics[width = 0.6\linewidth]{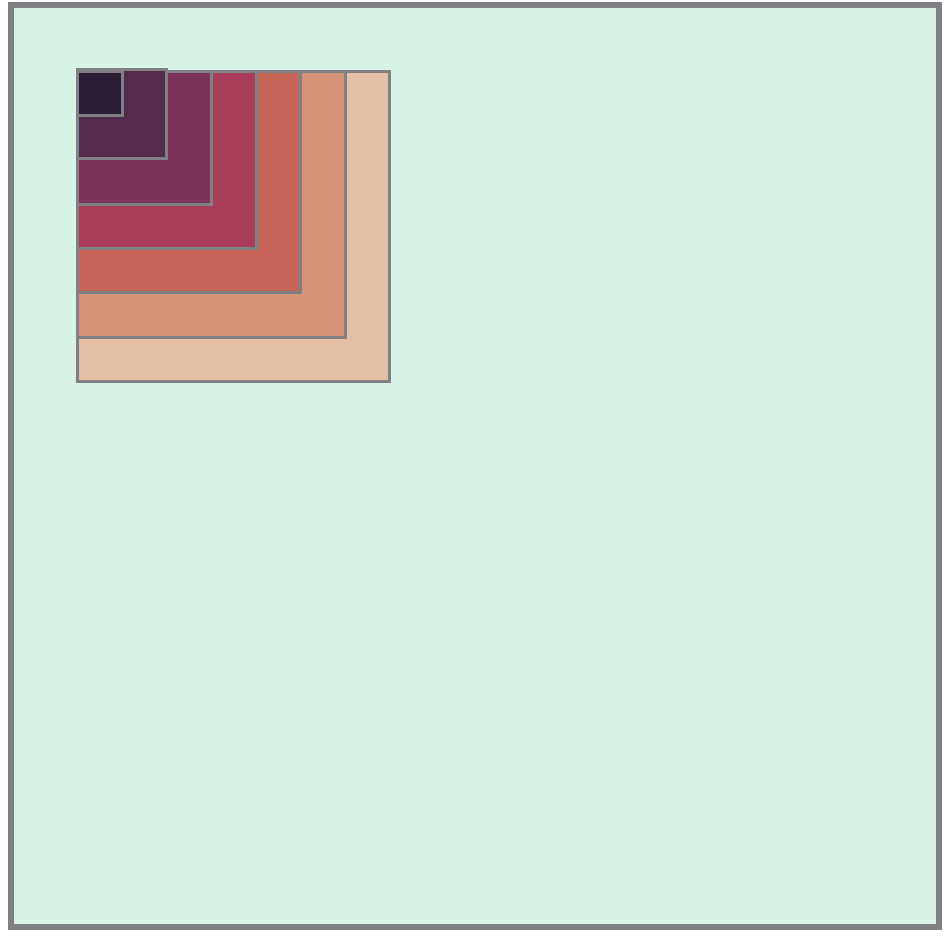}
    \caption{Visualization of different patch sizes compared to the input image. The smallest patch size is 20x20 pixels, the largest is 140x140 pixels.}
    \label{fig:patchsizerelative}
\end{figure}

Figure \ref{fig:patchsizevisual} demonstrates ASR for the evaluated patch sizes. A clear positive correlation between patch size and ASR can be seen. For patches smaller than 60x60 pixels we do not manage to achieve the ASR of more than 50\%. For the patch size of 100x100 pixels, the ASR is exceeds 90\% with even larger patches only leading to marginal improvements. Figure \ref{fig:patchsizerelative} further visualizes the coverage of each patch size with respect to the input image size.

\begin{figure}[t]
\centering
\begin{subfigure}[t]{0.485\linewidth}
    \includegraphics[width=1.0\textwidth]{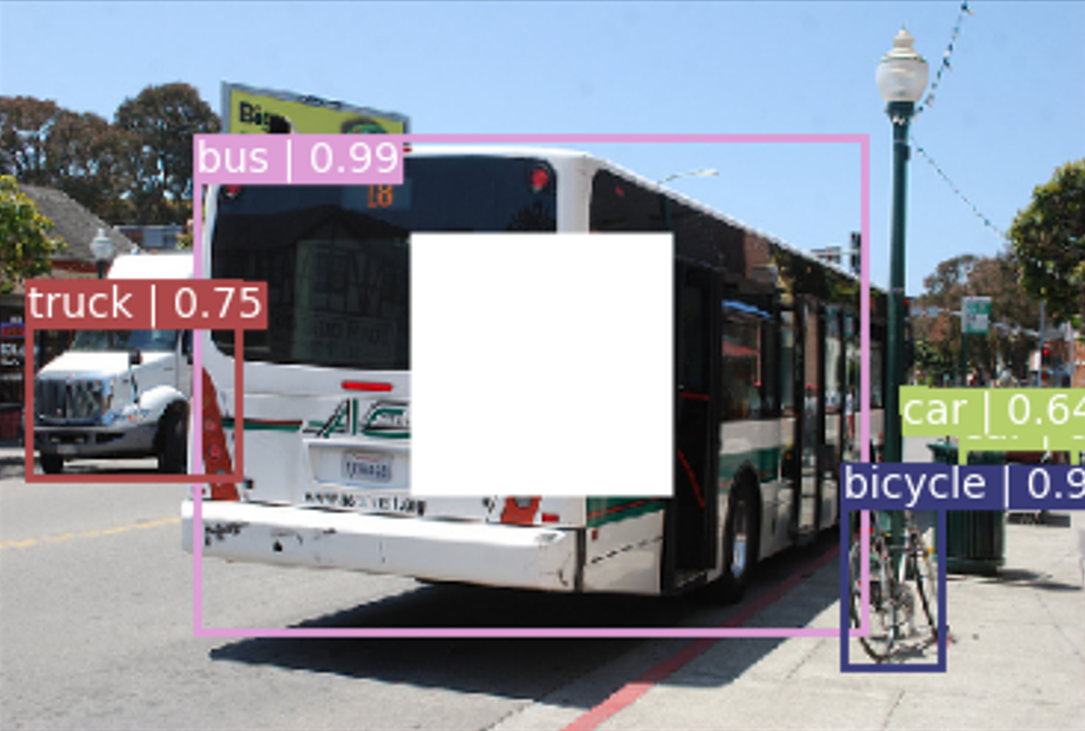}
  	\caption{White patch}
\end{subfigure}

\begin{subfigure}[t]{0.485\linewidth}
  \includegraphics[width=1.0\textwidth]{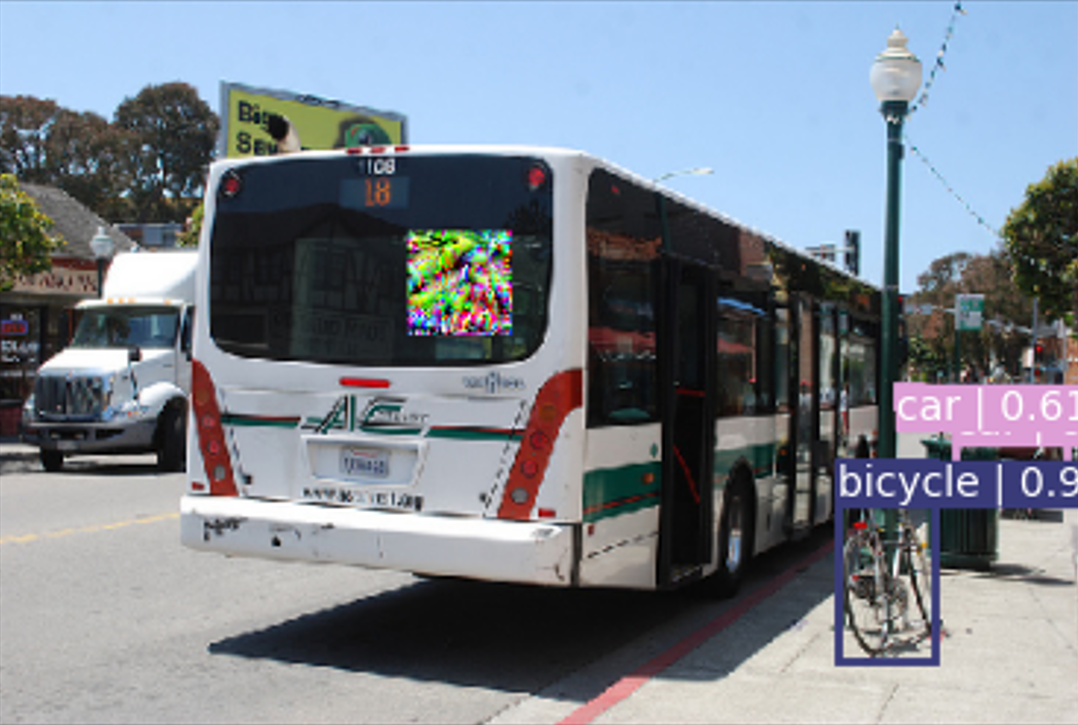}
  	\caption{Adversarial patch, 60x60 pixels}
\end{subfigure}
\begin{subfigure}[t]{0.485\linewidth}
  \includegraphics[width=1.0\textwidth]{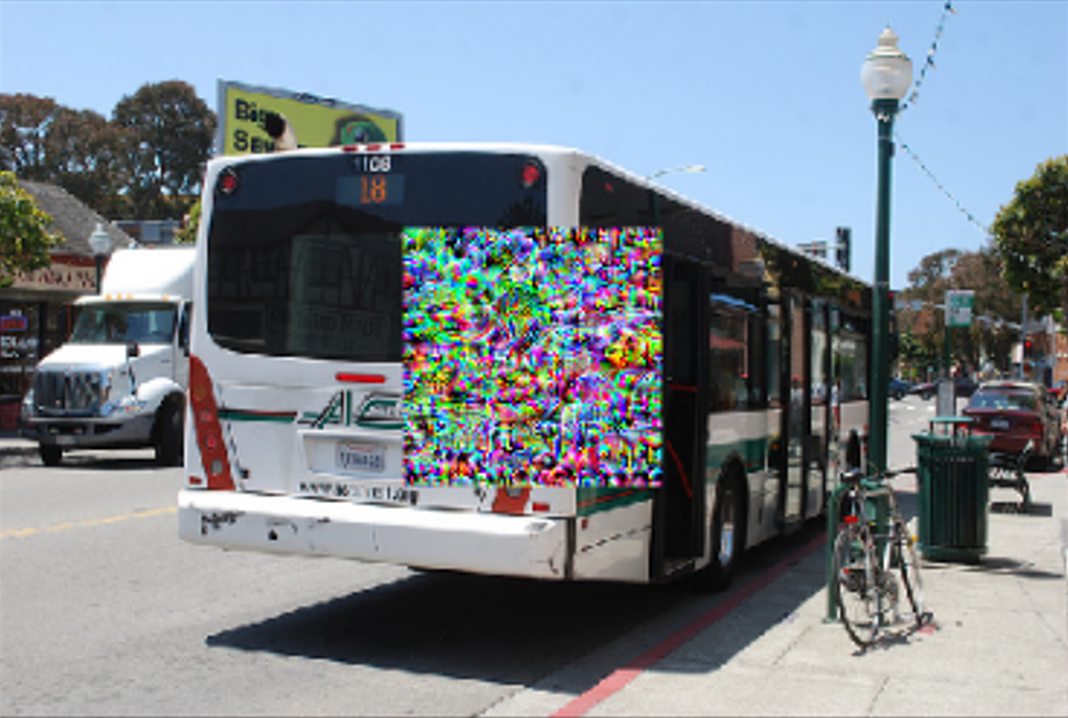}
  	\caption{Adversarial patch, 100x100 pixels}
\end{subfigure}
    \caption{Exemplary evaluation of patches of different size and comparison to a white benign patch, used to compute ASR.}
    \label{fig:patchsize_img}
\end{figure}

Based on this evaluation, we use patches of size 100x100 pixels in further experiments. This patch size leads to high ASR values of over 90\% under ideal conditions while being small enough with coverage of 5.78\%. Figure~\ref{fig:patchsize_img} shows exemplary evaluation of different patch sizes on a single image.

\subsection{Patch Initialization}\label{initialisationofpatch}
We evaluate five methods to initialize a patch: three unicolor approaches (black, white, and gray), noise generated with a uniform distribution, and noise generated with a normal distribution with $\mathcal{N} {(0.5,0.5)}$.

\begin{figure}[t]
    \centering
    \includegraphics[width = \linewidth]{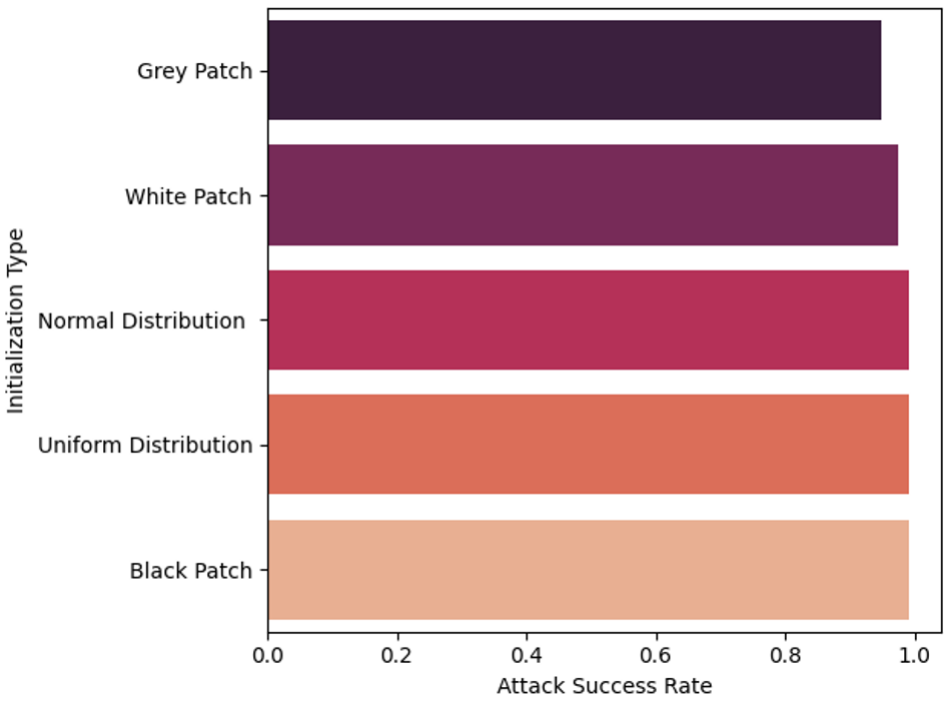}
    \caption{ASR for different patch initialization approaches.}
    \label{fig:initialisation}
\end{figure}

As Figure \ref{fig:initialisation} demonstrates, no clear advantage due to the usage of any initialization method is visible. We have observed, however, that patches initialized with a normal distribution converge faster. For this reason we further rely on this initialization method.

\subsection{Impact of Randomness}\label{randomeffects}
To better understand the impact of randomness in our setting, we started ten runs with the same parameters but different random seeds. For this, we used a constant data set of 10,000 images randomly selected from the COCO training data set and trained our patch for 100 epochs with a batch size of 15. The results are reported for the COCO validation dataset.
 
First, we analyzed the variation of the ASR within the last five epochs of each run. Within these, we found a maximum variation of 2.15\% in ASR. To compensate for this rather large variation and allow a comparison of different runs, we also used ASR averaged over five epochs of each run. The comparison has revealed a variation of the ASR of up to 3.75\%. Therefore, the reported ASR results can fluctuate within this range solely due to different random seeds.

\subsection{Patch Position}\label{positionofpatch}
Next, we evaluated the preferable method to place a patch during training with a goal to reach patch translation robustness. For this, we have studied three methods to position a patch during training: (1) placing a patch at a fixed position in the image center, (2) dynamic window approach, and (3) randomly placing a patch over the whole input image. 

The experiments were performed on a COCO subset containing 10,000 images, all patches were trained for 500 epochs. The results are reported for the COCO validation dataset.

Training a patch at a fixed position converged quickly and led to high ASR of 95\%, when the same position is used for evaluation. However, as soon as the patch position is only slightly altered e.g. for one pixel, ASR decreases substantially to around 18\%. Placing a patch at random positions, however, increased the ASR  significantly, reaching 72.1\%.

For the dynamic window approach, we have evaluated different strategies to initialize and increase the window size. In particular, a dynamic window that starts at the center and then increases in size each epoch, has demonstrated the best convergence. We have also evaluated temporary reductions in window size during training to ensure better patch performance on earlier training stages. However, overall, the random patch placing still led to stronger patch attacks. Interestingly, no significant difference in training time between random position and dynamic window could be observed (36 hours each). However, patches trained at a fixed position required only about 31 hours, making this method relatively faster.

Figure \ref{fig:heatmaps} demonstrates the heatmaps, showing the placement of a patch midpoint throughout the training process. The heatmap for the fixed patch position reveals that a patch is evidently not moving during the entire training. The heatmap for the dynamic window approach has an increasingly large window in which a patch was inserted. Finally, the heatmap for random positioning shows that all possible locations for a patch midpoint were used during the training with the equal frequency. Interestingly, as the patch position changes from fixed to random, the visual patterns in a patch become more sophisticated.   

Overall, we can conclude, that moving the patch randomly over the whole target area as early as possible seems to be a viable strategy to achieve high patch translation robustness.

\subsection{Intra-Batch Variation}\label{ibvar}
We further assess the impact of the variations in patch placement within a batch. In addition to changing patch position only after processing a batch, we have also evaluated varying patch position for each image in a batch with a goal of increasing translation robustness by optimizing a patch for multiple positions simultaneously for each gradient descent step.

For this, we conducted ten training runs with 10,000 training images and 100 epochs at a batch size of 15, five runs with identical positions, and five runs with random patch position in a batch. Table \ref{tab:ASR_comparison} summarizes the results.  Additional variation within a batch evidently improves ASR. Moreover, it also leads to a smoother loss convergence.

\begin{table}[t]
\caption{Comparison of mean ASR for different approaches to place a patch during training. IB stands for intra-batch variation}
\begin{center}
	\label{tab:ASR_comparison}
		\begin{tabular}{|l|c|c|c|c|c|}
			\hline
			 & \textbf{Fixed} & \multicolumn{2}{|c}{\textbf{Dynamic Window}} & \multicolumn{2}{|c|}{\textbf{Random}}  \\
			 &   & \textbf{no IB} & \textbf{with IB} & \textbf{no IB} & \textbf{with IB}  \\
			\hline
			Mean ASR& 18.2\% & 68.3\% & 69,2\% & 72.1\% & {\bf73.5\%} \\
			\hline
	\end{tabular}
\end{center}
\end{table}

\subsection{Experiments in the Real World}\label{Application in the Real World}

We have printed the patches, obtained in the previous experiments, and evaluated them in the real-world setting. As expected, since no patch transformations beside translation were involved in patch training, we have observed much lower ASR. No values beneath the confidence threshold of 0.5 could be reached, which would make objects vanish. 

To mitigate it, we have applied the EoT approach~\cite{athalye2018synthesizing}. In particular, we have introduced random hue and brightness changes, as well as random decreases in saturation and contrast. We also slightly rotated the patch by random degrees within $\pm15^\circ$ to account for a human user holding a patch. Furthermore, we randomly set patch size between 60 and 120 pixels to achieve scaling robustness and thus better represent the different distances between the patch and the camera.

\begin{figure*}[t!]
\begin{center}
\begin{tabular}{cccc}
& Fixed position (center) & Dynamic window approach & Random positioning \\

\rotatebox{90}{\hspace{5 mm} Heatmap} & \includegraphics[width=1.7in]{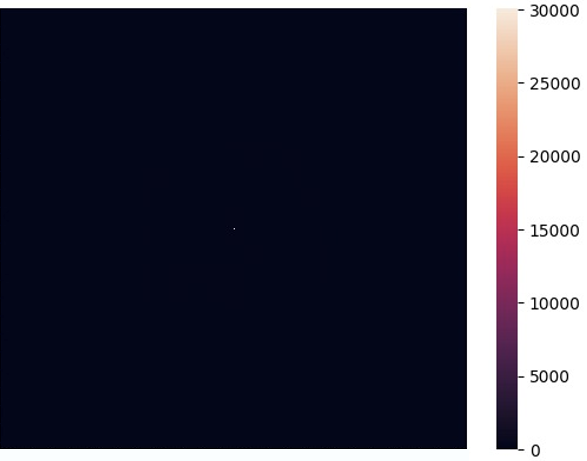} & 
  \includegraphics[width=1.7in]{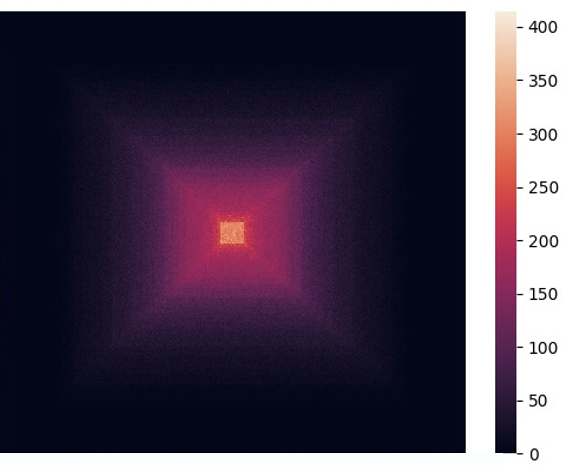} &   
  \includegraphics[width=1.7in]{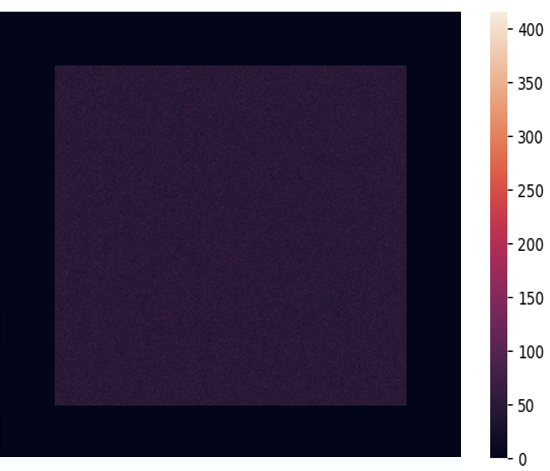} \\
  
\rotatebox{90}{\hspace{5 mm} Without augmentations} &  \includegraphics[width=1.7in]{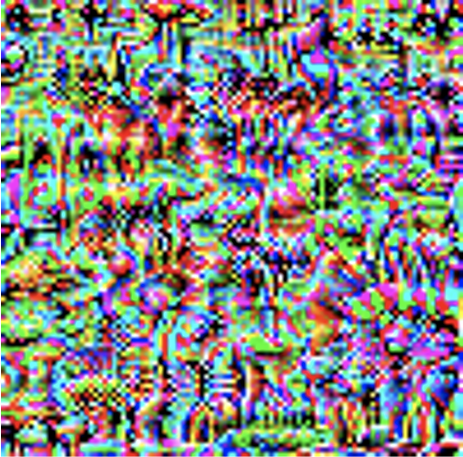} & 
    \includegraphics[width=1.7in]{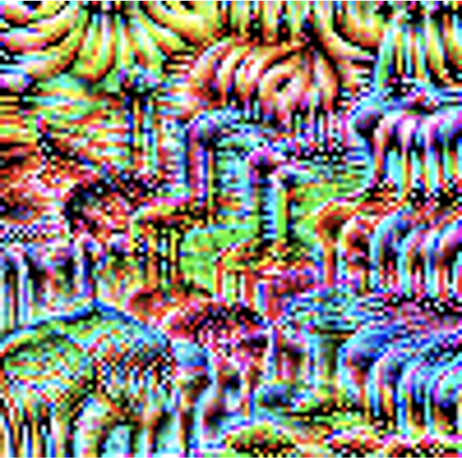} & 
    \includegraphics[width=1.7in]{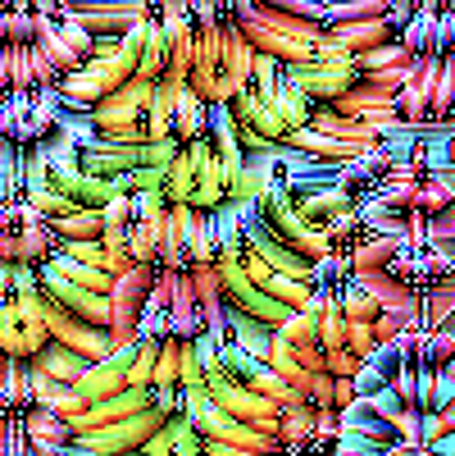} \\
    
\rotatebox{90}{\hspace{5 mm} With augmentations} &   & 
    \includegraphics[width=1.7in]{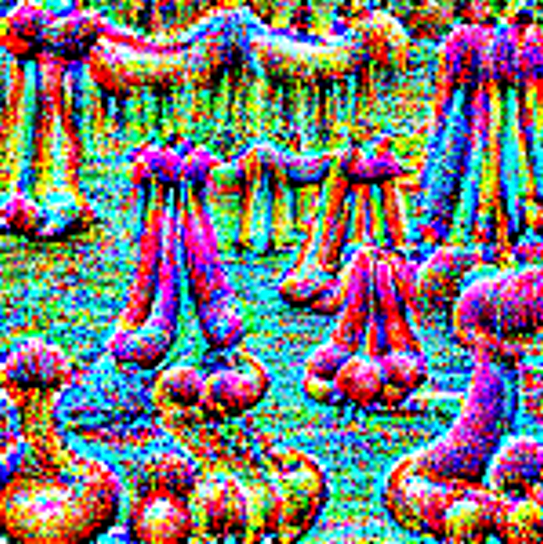} & 
    \includegraphics[width=1.7in]{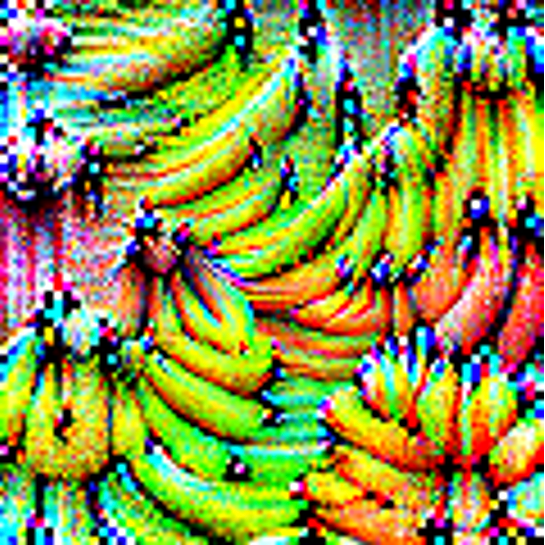} \\

\end{tabular}
\end{center}
\caption{Evaluation of three patch placement strategies.}
\label{fig:heatmaps}
\end{figure*}

We have performed five different runs with the formerly introduced data set of 10,000 COCO images and a batch size of 15 and reached the ASR averaging at 35.8\% for the dynamic window approach and 56.1\% for the random placement approach, which is significantly lower that in the digital setting. Interestingly, different patch placement strategies led to a significant difference in the learned visual patterns (see Figure \ref{fig:heatmaps}).

\section{Conclusion}\label{conclusionandprospects}
In this work, we have studied the impact of different hyperparameters during the generation of universal adversarial patches against object detection. Our goal was to completely suppress all detections in an input image. The attack was performed with PGD, a single patch was generated for the whole dataset in a universal manner. We have introduced the adversarial loss, which aims at minimizing the objectness scores and does not require ground truth labels. For evaluation, we have defined the attack success rate (ASR) metric, which involves comparison to a white benign patch, thus eliminating the influence of an overlap of an object with a patch.

We have analyzed different patch initialization methods, patch size, as well as patch placement strategies. Our approach has demonstrated a high attack success rate in controlled laboratory experiments, in most cases all object detections could be suppressed. 

Our experiments with three different patch placement strategies during training have demonstrated, that increased variation of patch position during training leads to the enhanced translation robustness of the patch. The described dynamic window approach has shown no benefit in terms of attack strength or training time. We have also determined, that in order to obtain a stronger attack, patch position should vary as much as possible between training epochs as well as within batches. The strongest attack could be achieved when inserting patches randomly over the epochs and also within every batch, which yielded an ASR of 73.5\%. 

Finally, to transfer the attack to the real world, we have introduced different transformations to the patch training process to ensure rotations and scaling robustness, as well as account for differences in brightness and contrast. As a result, in case of random patch positioning,  we could reach ASR of 56.1\% which led to object vanishing.

\section*{Acknowledgment}

The research leading to these results is funded by the German Federal Ministry for Economic Affairs and Climate Action within the project “KI Absicherung“ (grant 19A19005W) and by KASTEL Security Research Labs. The authors would like to thank the consortium for the successful cooperation

\bibliographystyle{IEEEtran}
\bibliography{references.bib}

\end{document}